\documentclass[runningheads]{llncs}
\usepackage{graphicx}
\usepackage{latexsym}
\usepackage{soul}
\usepackage{url}
\usepackage{graphicx}
\usepackage{amsmath}

\usepackage{amsthm}
\usepackage{amssymb}
\usepackage{booktabs}
\usepackage[ruled,linesnumbered]{algorithm2e}
\usepackage{multirow}
\usepackage{diagbox}
\usepackage{subfigure}
\usepackage{float}
\usepackage{fancyhdr}
\usepackage{makecell}
\usepackage{hyperref}
\newtheorem{notation}{Notation}
\usepackage[bottom]{footmisc}
\usepackage{caption}
\usepackage[T1]{fontenc}
\usepackage{graphicx}
\begin{document}

\title{Incorporating Higher-order Structural Information for Graph Clustering}


\author{%
Qiankun Li$^{*}$ \and
Haobing Liu\thanks{These authors contributed equally to this work and are co-first authors.} \and
Ruobing Jiang\thanks{Corresponding author.} \and
Tingting Wang}

\authorrunning{Li et al.}

\institute{
Department of Computer Science and Technology, Ocean University of China, Qingdao, China}


\maketitle

\begin{center}
\email
{abcliqiankun@stu.ouc.edu.cn,
haobingliu@ouc.edu.cn,\\
jrb@ouc.edu.cn,
wtt2022@stu.ouc.edu.cn}
\end{center}

\begin{abstract}
Clustering holds profound significance in data mining. In recent years, graph convolutional network (GCN) has emerged as a powerful tool for deep clustering, integrating both graph structural information and node attributes. However, most existing methods ignore the higher-order structural information of the graph. Evidently, nodes within the same cluster can establish distant connections. Besides, recent deep clustering methods usually apply a self-supervised module to monitor the training process of their model, focusing solely on node attributes without paying attention to graph structure. 
In this paper, we propose a novel graph clustering network to make full use of graph structural information. To capture the higher-order structural information, we design a graph mutual infomax module, effectively maximizing mutual information between graph-level and node-level representations, and design a trinary self-supervised module that includes modularity as a structural constraint.
Our proposed model outperforms many state-of-the-art methods on various datasets, demonstrating its superiority.
\end{abstract}




\section{Introduction}

Clustering plays a pivotal role in data mining, facilitating the categorization of real-world objects into clusters based on attributes or structural information in an unsupervised way. Its practical applications are diverse. For instance, clustering users and products offers valuable insights for recommender platforms~\cite{Experience-Aware2015}. In biological networks~\cite{CarloNicolini2016CommunityDI}, clustering contributes to unveiling intricate mechanisms in the human body. The application of clustering in social networks aids in the identification of terrorist organizations~\cite{FirasSaidi2018ANA}, contributing to the maintenance of regional security.

Clustering methods have evolved for a long time, aiming to incorporate richer information. Initially, methods focused on either graph structure~\cite{METIS} or node attributes~\cite{DEC}. However, researchers later realized that achieving superior clustering performance requires integrating both node attributes and graph structure, crucial for capturing node similarities. As deep learning expands its applications across various domains~\cite{TBD}, clustering methods can leverage its ability to utilize information from both sides~\cite{Survey}. Notably, approaches based on graph convolutional networks (GCNs)~\cite{MHGCN,GCN2Y} aggregate neighbors' features to obtain node representations and conduct clustering, resulting in more similar representations for nodes connected by shorter paths.

However, most existing methods are limited to handling higher-order structural information. For instance, in the Citeseer dataset, 52.64\% of nodes within the same cluster are connected through 7 or more hops. Merely employing GCN fails to capture this complex structural information. Thus, the challenge lies in \textit{how to capture the higher-order structural information of the graph}. Additionally, some methods combine GCN with autoencoder (AE) and employ a dual self-supervised module for training~\cite{DeyuBo2020StructuralDC,GuangyuHuo2021CaEGCNCF}. These methods generate soft clustering assignments by measuring node distances to cluster centers derived from the encoder. However, the target distribution emphasizes node attributes, neglecting graph structure. Hence, a key challenge is \textit{how to utilize graph structure to supervise the target distribution}.

In this paper, we propose a novel model that fully leverages the higher-order structural information of the graph and node attributes. Initially, an attribute-enriched GCN (AGCN) layer is constructed using a combination of encoder and GCN to aggregate neighbors' information. To overcome the first challenge, a graph mutual infomax module is designed to capture the higher-order structural information of the graph. To overcome the second challenge, a trinary self-supervised module is proposed to optimize the entire model, taking both graph structure and node attributes into account. The contributions of the paper are summarized as follows:
\begin{itemize}
    \item We propose a novel unsupervised deep clustering method, which is a \underline{h}igh\underline{er}-\underline{o}rder \underline{g}raph \underline{c}lustering \underline{n}etwork (HeroGCN). Our method fully considers \textit{the higher-order graph structural information} and node attributes for clustering.
    \item We allow our model to learn node representations guided by graph mutual information using graph contrast learning to capture the \textit{higher-order structural information} of the graph.
    \item We introduce modularity into the dual self-supervised module and propose a trinary self-supervised module to simultaneously consider both node attributes and \textit{graph structure} in training the model.
\end{itemize}

\section{Related Work}\label{sec:Related Work}
Both graph structure and node attributes are important for clustering. According to the information considered by clustering methods, these methods can be classified into two categories: 1) methods modeling graph structure or node attributes; 2) methods jointly modeling graph structure and node attributes.
\\\\
\textbf{Methods Modeling Graph Structure or Node Attributes.}
Early methods just model graph structure or node attributes for clustering. Infomap~\cite{Infomap} identifies cohesive and well-connected groups of nodes based on information-theoretic principles. DEC~\cite{DEC} derives clustering results from representations learned via an autoencoder. IDEC~\cite{XifengGuo2017ImprovedDE} keeps the decoder which is discarded in DEC and achieves better performance. These methods only utilize graph structure or node attributes for clustering, without considering them both.
\\
\\
\textbf{Methods Jointly Modeling Graph Structure and Node Attributes.}
Recent research has recognized the importance of considering both graph structure and node attributes. DAEGC~\cite{ChunWang2019AttributedGC} employs a graph attentional autoencoder to capture neighbor features of nodes. SDCN~\cite{DeyuBo2020StructuralDC} integrates GCN and AE with a dual self-supervised module. Further, CaEGCN~\cite{GuangyuHuo2021CaEGCNCF} proposes a cross-attention fusion module, merging GCN and autoencoder to enhance crucial information. These methods consider both structural information and node attributes, leading to improved clustering performance.

\section{Preliminaries}\label{sec:Preliminaries}
In this section, we introduce some basic definitions and some knowledge about autoencoder.

\begin{notation}[Attributed Graph]
  An attributed graph is defined as $G=(V, E, X)$. $V=\{v_1,v_2,...,v_n\}$ is the set of nodes, where $\lvert V \rvert=n$. $E$ is the set of edges, which can be represented by the adjacency matrix $A$. If there is an edge between $v_i$ and $v_j$, $A_{ij}=1$, otherwise $A_{ij}=0$. $X=\{x_1,x_2, ...,x_n\}$ is the attribute matrix, where $x_i$ is the attribute vector corresponding to node $v_i$.
\end{notation}

\noindent\textbf{Problem Formulation}\quad Given an attributed graph $G$ and the number of clusters $K$, the objective of clustering is to find a function $f:V\to C$ that assigns nodes into $K$ distinct clusters. $C=\{C_1, C_2, ..., C_K\}$, each cluster $C_k$ is a division of the graph $G$. $\forall{k, k'}$, $C_k\cap C_{k'}=\emptyset$. Nodes satisfying $f(v_i)=C_k$ belong to the $k$-th cluster. The clustering approach should ensure that nodes within the same cluster exhibit higher similarity in representations or stronger connections. From the perspective of graph structure, intra-cluster edges should outnumber inter-cluster edges.

\section{The Proposed Model}\label{sec:Method}
\begin{figure}[t!]
\setlength{\abovecaptionskip}{-0.1cm}
    \vspace{-0.4cm}
    \centering
    \includegraphics[scale=0.29]{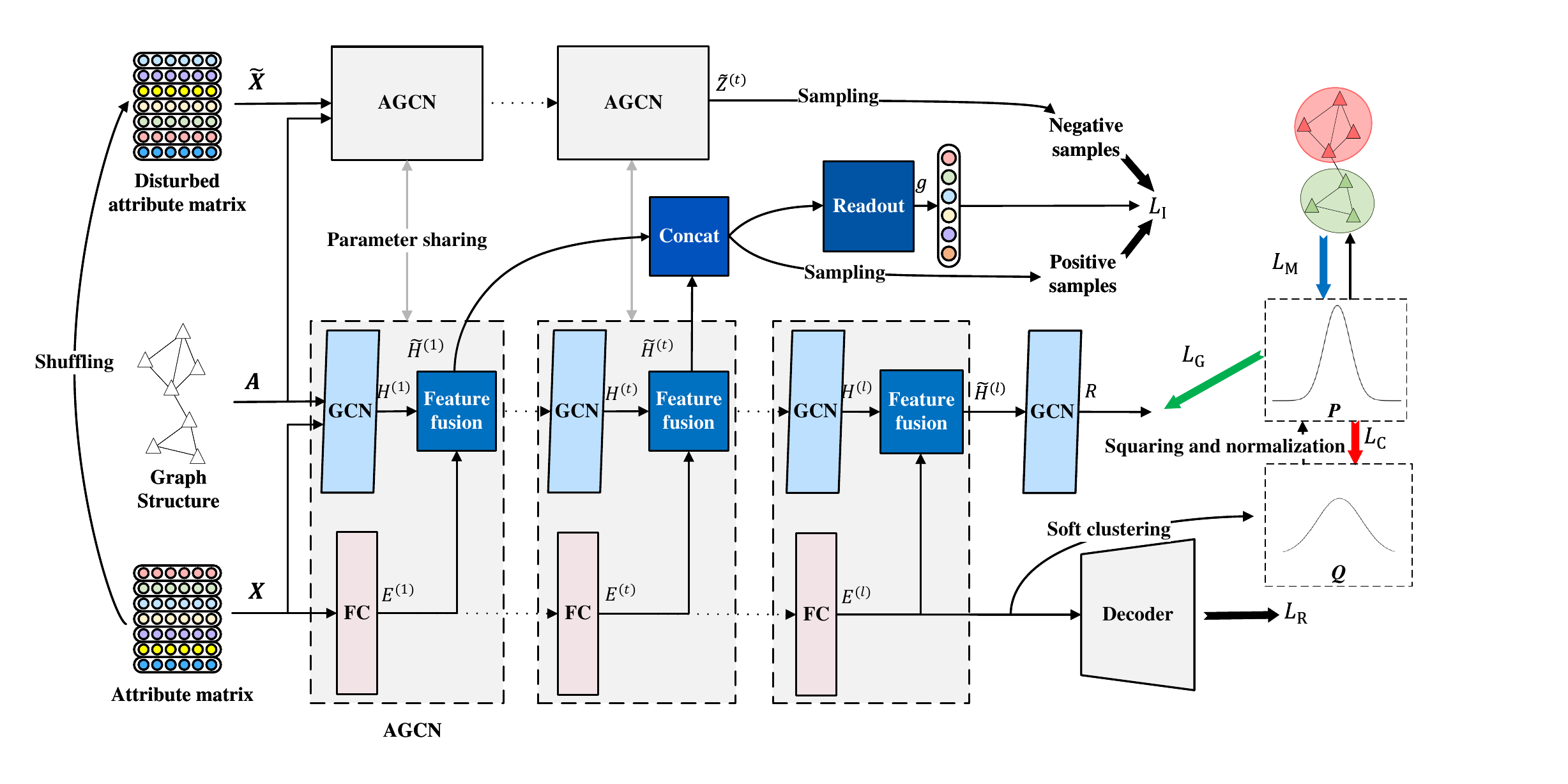}
    \caption{The framework of the proposed method HeroGCN.}\label{fig:my_label}
    \vspace{-0.5cm}
\end{figure}

In this section, we introduce our proposed model, HeroGCN, as depicted in Fig. \ref{fig:my_label}. Our model comprises three components: attribute-enriched GCN, graph mutual infomax module, and trinary self-supervised module.

\subsection{Attribute-enriched GCN}
In the AGCN layer, we employ fully connected layers to obtain effective representations by reconstructing the input from the representations. The network consists of two parts: the encoder and the decoder. Assuming that the encoder has $L$ layers, $l$ represents the layer number corresponding to the $l$-th layer encoder. The representations of the ${l}$-th layer encoder can be written as follows:
\begin{equation}\label{aeenc}
    \begin{aligned}
    &E^{(1)} = \sigma(W_{\rm{E}}^{(1)} X+b_{\rm{E}}^{(1)}),\\
    &E^{(l)} = \sigma(W_{\rm{E}}^{(l)} E^{(l-1)}+b_{\rm{E}}^{(l)}), \quad 2\leq l \leq L,
    \end{aligned}
\end{equation}
where $\sigma$ is the Relu function, $W_{\rm{E}}^{(l)}$ is the weight matrix of the $l$-th layer encoder, and $b_{\rm{E}}^{(l)}$ is the bias of the $l$-th layer encoder.

The decoder is the mirror of the encoder, which reconstructs the input from the representations learned by the encoder. The loss function is defined by comparing the input $X$ with the reconstructed input $\hat{X}$:
\begin{align}\label{loss_res} 
    L_{\rm{R}} = \frac{1}{2N} \sum_{i=1}^N\|x_i-\hat{x}_i\|_2^2 = \frac{1}{2N} \|X-\hat{X}\|_F^2, 
\end{align}
where $N$ is the number of samples. 

At the same time, we incorporate the node representations learned by the encoder into GCN to obtain more latent information. The convolution operation of GCN can be represented as follows:
\begin{equation}\label{agcnenc}
   \begin{aligned}
    &H^{(1)} = \sigma(\hat{A} X W^{(1)}),\\
    &H^{(l)} = \sigma(\hat{A} \tilde{H}^{(l-1)} W^{(l)}), \quad 2\leq l\leq L,
\end{aligned}%
\end{equation}
where $\sigma$ is the Relu function. $\hat{A}=\tilde{D}^{-\frac{1}{2}} \tilde{A}\tilde{D}^{-\frac{1}{2}}$, which is the symmetric normalization of the adjacency matrix. $\tilde{A}=A+I$, which is the self-connected adjacency matrix. $\tilde{D}$ is the degree matrix satisfying $\tilde{D}_{ii}=\sum_{j=1}^N\tilde{A}_{ij}$. $X$ is the attribute matrix and $W^{(l)}$ is the trainable weight matrix. $H^{(l)}$ is the node representations learned by the $l$-th GCN layer, which is obtained by doing convolution on the hybrid representations $\tilde{H}^{(l-1)}$. 
We can also get node representations $E^{(l)}$ at the $l$-th FC layer according to Eq.~\eqref{aeenc}. 

Then, we fuse node representations $H^{(l)}$ got from GCN with node representations $E^{(l)}$ got from antoencoder as:
\begin{align}\label{feature fusion}
    \tilde{H}^{(l)}=\alpha H^{(l)}+(1-\alpha)E^{(l)}, \quad 1\leq l\leq L,
\end{align}%
where $\alpha$ is the fusion coefficient. In the subsequent modules, we can use such hybrid representations to further optimize the training process of our model.

\subsection{Graph Mutual Infomax Module}
To capture the higher-order structural information while clustering, HeroGCN integrates a graph mutual infomax module, which maximizes the mutual information between graph-level and node-level representations.

For the node-level representations, the raw data of node attributes is shuffled and fed into AGCN to get the output $\tilde{Z}$. $\tilde{H}$ is the output of AGCN with the original data at the same layer. Then we concatenate the outputs of the first $t$ AGCN layers to get positive samples, which can be described as:
\begin{align}\label{samples}
    \hat{h}_i=\tau(\tilde{h}_i^{(1)}, \tilde{h}_i^{(2)},  ...  , \tilde{h}_i^{(t)}),
\end{align}%
where $\tau$ represents the concatenation of vectors, and we get the positive samples $\hat{H} = \{\hat{h}_1, \hat{h}_2, ...,  \hat{h}_n\}$. The negative samples are obtained in the same way. For the graph-level representation, all the positive samples are summarized into a vector. A readout function is designed to get the graph-level representation, which can be described as: 
\begin{align}\label{graph representation}
    g=\frac{1}{N} \sum_{i=1}^N\hat{h}_i,
\end{align}%
where $N$ is the number of nodes and the graph-level representation $g$ is a summary of all positive samples. 

A standard binary cross-entropy (BCE) loss is employed to compare the positive and negative samples. The graph mutual information can be maximized with the following loss function:
\begin{align}\label{loss_gmi}
    L_{\rm{I}}=-\frac{1}{N+M} (\sum_{i=1}^N\mathbb{E}_{(X,A)}[{\rm log} D(\hat{h}_i, g)]+\sum_{j=1}^M\mathbb{E}_{(\tilde{X},A)}[{\rm log}
    (1-D(\hat{z}_i, g))]),
\end{align}%
where $N$ and $M$ denote the number of positive and negative samples.
$D$ is the discriminator that uses a bilinear function to calculate the patch-summary score, which can be described as:
\begin{align}
    D(\hat{h}_i, g)=\sigma(\hat{h}_i W_{\rm{S}} g),
\end{align}%
where $\sigma$ is the sigmoid function and $W_{\rm{S}}$ is a trainable scoring matrix. 

\subsection{Trinary Self-supervised Module}
 The dual self-supervised module is widely adopted in clustering research~\cite{DeyuBo2020StructuralDC,GuangyuHuo2021CaEGCNCF}. It utilizes a highly confident target distribution to supervise model training. However, this target distribution is solely obtained from the autoencoder, overlooking the importance of graph structural information in guiding clustering. To address this issue, our model introduces a trinary self-supervised module, incorporating modularity to monitor the target distribution based on graph structure.

First, we use the Students' $t$-distribution~\cite{LaurensvanderMaaten2008VisualizingDU} to obtain the clustering assignments, which can be described as follows:
\begin{align}\label{Q distribution}
    q_{ik}=\frac{(1+\|e_i-\mu_k\|^2)^{-1}}{\sum_{k'=1}^K(1+\|e_i-\mu_{k'}\|^2)^{-1}},
\end{align}%
where $e_i$ is the representation of the $i$-th node learned by the encoder, $\mu_k$ is the representation of the $k$-th cluster center initialized by the pre-trained encoder and is trainable, $q_{ik}$ is the probability of assigning node $i$ to cluster $k$, $K$ is the number of clusters. The overall soft clustering assignments $Q=[q_{ik}]$.

Second, with the soft clustering assignments $Q$, the target distribution can be calculated with the following function:
\begin{align}\label{P distribution}
    p_{ik}=\frac{q_{ik}^2 /\sum_{i=1}^Nq_{ik}}{\sum_{k'=1}^K(q_{ik'}^2 /\sum_{i=1}^Nq_{ik'})},
\end{align}%
which has higher confidence compared with the original distribution. 

Next, the target distribution is used to supervise the original distribution. A KL divergence loss between $Q$ and $P$ is applied as follows:
\begin{align}\label{loss_PQ}
    L_{\rm{C}}={\rm KL}(P\|Q)=\sum_{i=1}^N\sum_{k=1}^Kp_{ik} {\rm log}\frac{p_{ik}}{q_{ik}},
\end{align}%
where $N$ is the number of nodes. The loss function encourages node representations to become more similar to cluster centers, serving as the first self-supervised mechanism for the encoder. Hybrid node representations from AGCN undergo another convolution operation for clustering, outlined as follows::
\begin{align}\label{R distribution}
    R={\rm softmax}(\hat{A} \tilde{H} W),
\end{align}%
so that we can get clustering assignments $R$, which is supervised by the target distribution $P$ through the following loss function:
\begin{align}\label{loss_PR}
    L_{\rm{G}}={\rm KL}(P\|R)=\sum_{i=1}^N\sum_{k=1}^Kp_{ik} {\rm log}\frac{p_{ik}}{r_{ik}},
\end{align}%
which serves as the second self-supervised mechanism in our model.

Finally, we incorporate modularity to monitor the target distribution on graph structure, serving as another self-supervised mechanism from graph structure. Higher modularity indicates more intra-cluster edges and fewer inter-cluster edges in the graph. The loss function can be expressed as follows:
\begin{align}\label{loss_mod}
    L_{\rm{M}}=-\frac{1}{2m} \sum_{ij}\sum_{k=1}^K(A_{ij}-\frac{d_i d_j}{2m}) p_{ik} p_{jk}, 
\end{align}%
where $m$ is the number of edges, $A_{ij}$ is the element of adjacency matrix, $d_i$ is the degree of node $i$, $p_{ik}$ is the probability of assigning node $i$ to cluster $k$ in the distribution $P$. Now the trinary self-supervise module has formed, which takes both graph structure and node attributes into account. 

\subsection{Loss Function and Training Process}
The loss function of our model can be described as follows:
\begin{align}\label{loss_overall}
    L=L_{\rm{R}}+\lambda_1 L_{\rm{I}}+\lambda_2 L_{\rm{C}}+\lambda_3 L_{\rm{G}}+\lambda_4 L_{\rm{M}},
\end{align}%
where $\lambda_1$, $\lambda_2$, $\lambda_3$, $\lambda_4$ are the coefficients that control the balance of the five items in the loss function.

The final assignment $y_i$ is determined by assigning node $i$ to the cluster $k$ with the highest probability in the distribution $R$:
\begin{align}\label{results}
    y_i = \arg\max_kr_{ik}.
\end{align}%

\section{Experiments}\label{sec:Experiments}
Experiments have been conducted to demonstrate the performance of HeroGCN from different aspects.

\subsection{Datasets}
The HeroGCN proposed in this paper is evaluated on five widely used datasets, namely ACM\footnote{http://dl.acm.org/}, DBLP\footnote{https://dblp.uni-trier.de}, Citeseer\footnote{http://citeseerx.ist.psu.edu/index}, USPS, and HHAR. The last two datasets are generated the same as SDCN~\cite{DeyuBo2020StructuralDC}. The detailed descriptions are presented in Table \ref{tab:datasets}.

\subsection{Baselines}
We compare our model with several state-of-the-art clustering methods. Baselines are classified into two categories: methods modeling either graph structure or node attributes, including METIS~\cite{METIS}, K-Means~\cite{JAHartigan1979AKC}, AE~\cite{GeoffreyEHinton2006ReducingTD}, and IDEC~\cite{XifengGuo2017ImprovedDE}; and methods jointly modeling graph structure and node attributes, including VGAE~\cite{ThomasKipf2016VariationalGA}, DAEGC~\cite{ChunWang2019AttributedGC}, ARGA~\cite{ShiruiPan2020LearningGE}, ProGCL~\cite{ProGCL}, SDCN~\cite{DeyuBo2020StructuralDC}, and CaEGCN~\cite{GuangyuHuo2021CaEGCNCF}.\\

\subsection{Implementation Details}
We evaluate our model's performance using four standard metrics: Accuracy (ACC), Normalized Mutual Information (NMI), Average Rand Index (ARI), and macro F1-score (F1). Higher scores indicate superior performance. We compare against baselines following original settings and conduct 10 repetitions of experiments with HeroGCN with different random seeds.

HeroGCN is implemented using PyTorch 1.9.1, with hyperparameters tuned through grid search. The AGCN outputs are structured as 500-500-2000-10 dimensions, aligned with the pre-trained autoencoder settings from SDCN\cite{DeyuBo2020StructuralDC}. We train our model for 1500 epochs with Adam optimizer. The sampled layer number $t$ is set to 3 and the fusion coefficient is set to 0.5. The hyperparameters \{$\lambda_1, \lambda_2, \lambda_3, \lambda_4$\} in the loss function are adjusted to \{0.5, 0.1, 0.01, 0.05\} for balance. The batch size is set to 256, with varying learning rates: $1\times10^{-4}$ for ACM and USPS, $3\times10^{-4}$ for DBLP, $2\times10^{-4}$ for Citeseer, and $5\times10^{-5}$ for HHAR.

\begin{table}[b!]
\vspace{-0.8cm}
\tiny
\centering
\caption{Clustering results on benchmark datasets. The best results and the second-best results are highlighted. $\star$ manifests that HeroGCN's result is distinct from the best baseline with a significance level $p$-value $\textless$ 0.05.}
\begin{tabular}{c|c|cccc|ccccccc}
\hline
Dataset  & \diagbox{Metric}{Method} & METIS & K-Means & AE & IDEC & VGAE & ARGA & DAEGC & SDCN & ProGCL & CaEGCN & HeroGCN \\
\hline
\multirow{4}{*}{ACM} & ACC & 33.98 & 67.31 & 81.83 & 86.45 & 84.13 & 83.27 & 86.94 & 89.79 & 86.98 & \underline{90.12} & \bf{91.07}$^{\star}$ \\
& NMI & 0.02 & 32.44 & 49.30 & 58.24 & 53.20 & 50.39 & 56.18 & 66.93 & 59.13 & \underline{67.03} & \bf{70.39}$^{\star}$\\
& ARI & -0.04 & 30.60 & 54.64 & 64.21 & 57.72 & 56.46 & 59.35 & 72.32 & 65.22 & \underline{73.00} & \bf{75.54}$^{\star}$\\
& F1 & 33.98 & 67.57 & 82.01 & 86.32 & 84.17 & 83.35 & 87.07 & 89.77 & 87.03 & \underline{90.09} & \bf{91.04}$^{\star}$\\
\hline
\multirow{4}{*}{DBLP} & ACC & 26.57 & 38.65 & 51.43 & 65.71 & 58.59 & 54.50 & 62.05 & 66.48 & \underline{71.06} & 68.23 & $\rm\textbf{72.62}^{\star}$\\
& NMI & 0.12 & 11.45 & 25.40 & 30.80 & 26.92 & 20.19 & 32.49 & 32.92 & \underline{38.03} & 33.88 & \bf{39.15}$^{\star}$\\
& ARI & 0.02 & 6.97 & 12.21 & 32.10 & 17.92 & 19.49 & 21.03 & 33.54 & \underline{37.20} & 36.17 & \bf{41.57}$^{\star}$\\
& F1 & 26.34 & 31.92 & 32.53 & 64.39 & 58.69 & 53.43 & 61.75 & 65.43 & \underline{70.42} & 66.69 & \bf{72.21}$^{\star}$\\
\hline
\multirow{4}{*}{Citeseer} & ACC & 18.21 & 39.32 & 57.08 & 60.23 & 60.97 & 59.12 & 64.54 & 67.21 & 64.95 & \underline{68.02} & \bf{70.33}$^{\star}$\\
& NMI & 0.19 & 16.94 & 27.64 & 30.74 & 32.69 & 30.69 & 36.41 & 39.44 & 39.45 & \underline{40.00} & \bf{43.05}$^{\star}$\\
& ARI & -0.03 & 13.43 & 29.31 & 29.24 & 33.13 & 31.38 & 37.78 & 41.68 & 38.83 & \underline{42.40} & \bf{45.57}$^{\star}$\\
& F1 & 17.92 & 36.08 & 53.80 & 52.30 & 57.70 & 54.85 & \underline{62.20} & 60.92 & 60.30 & 61.38 & \rm\bf{62.41}$^{\star}$\\
\hline
\multirow{4}{*}{USPS} & ACC & 16.76 & 66.82 & 44.02 & 76.84 & 63.81 & 71.96 & 73.55 & 77.22 & 73.73 & \underline{77.55} & \bf{80.15}$^{\star}$\\
& NMI & 0.20 & 62.72 & 48.50 & 77.95 & 70.04 & 68.59 & 71.12 & 79.07 & 73.25 & \underline{79.23} & \bf{80.59}$^{\star}$\\
& ARI & 0.00 & 54.64 & 30.82 & 70.11 & 56.36 & 60.81 & 63.33 & \underline{71.10} & 64.32 & 71.07 & \bf{73.49}$^{\star}$\\
& F1 & 3.00 & 64.94 & 36.65 & 75.65 & 58.61 & 70.93 & 72.45 & 76.26 & 71.51 & \underline{76.34} & \rm\bf{78.17}$^{\star}$\\
\hline
\multirow{4}{*}{HHAR} & ACC & 17.76 & 59.98 & 46.21 & 79.20 & 62.52 & 70.40 & 76.51 & 84.49 & 61.90 & \underline{87.42} & \bf{88.21}$^{\star}$\\
& NMI & 0.07 & 58.87 & 36.10 & 79.60 & 60.59 & 71.54 & 69.10 & 80.21 & 67.26 & \bf{82.56} & \underline{81.44}\\
& ARI & 0.01 & 46.09 & 22.57 & 70.33 & 46.01 & 61.14 & 60.38 & 72.92 & 52.81 & \underline{76.27} & \bf{77.00}$^{\star}$\\
& F1 & 17.67 & 58.33 & 41.82 & 73.33 & 56.96 & 66.67 & 76.89 & 82.97 & 59.82 & \underline{87.24} & \rm\bf{88.06}$^{\star}$\\
\hline
\end{tabular}
\label{tab:baselines}
\vspace{-0.4cm}
\end{table}

\subsection{Experiment Results}
Table \ref{tab:baselines} presents the clustering results of our model and baselines on five datasets. HeroGCN achieves the best or the second-best results on all evaluation matrices. Taking DBLP for example, HeroGCN boosts accuracy by 6.43\% over CaEGCN and 9.24\% over SDCN. The NMI increases by 15.55\% over CaEGCN and 18.92\% over SDCN. The ARI rises by 14.93\% over CaEGCN and 23.94\% over SDCN. The F1-score elevates by 8.28\% over CaEGCN and 10.36\% over SDCN.

Our experiments reveal that some methods considering both graph structure and node attributes underperform compared to IDEC, which exclusively focuses on node attributes. The observed discrepancy is ascribed to the intrinsic over-smoothing issue within GCN, resulting in similar representations. This problem can be alleviated by combining GCN and encoder, as exemplified by the notable performance distinction between DAEGC and SDCN.

We observe that IDEC, DAEGC, SDCN, and CaEGCN outperform other clustering methods. A key factor contributing to their success is the inclusion of a self-supervised mechanism, facilitating the model in learning cluster-oriented representations, which is also integrated and enhanced in our model. 

\subsection{Ablation Study}
We conducted an ablation study on benchmark datasets. Accuracy is used to verify the effectiveness of different components of our model. The variants of HeroGCN are defined as follows:

\begin{figure}[t!]
\vspace{-0.4cm}
 \begin{minipage}{0.4\textwidth} 
 \centering 
  \captionof{table}{Benchmark Datasets}
  \scriptsize
\begin{tabular}{lrrcr}
\hline
Dataset  & \#Nodes & \#Edges & \#Classes & \#Attributes \\
\hline
ACM       & 3,025  & 13,128 & 3 & 1,870    \\
DBLP       & 4,058  & 3,528  & 4 & 334   \\
Citeseer    & 3,327  & 4,732  & 6 & 3,703  \\
USPS        & 9,298   & 21,452 & 10 & 256   \\
HHAR        &  10,299    & 38,039 & 6 & 561 \\
\hline
\end{tabular}
\label{tab:datasets}
 \end{minipage} 
  \begin{minipage}{0.75\textwidth} 
  \setlength{\abovecaptionskip}{0.1cm}
 \centering 
 \includegraphics[scale=0.35]{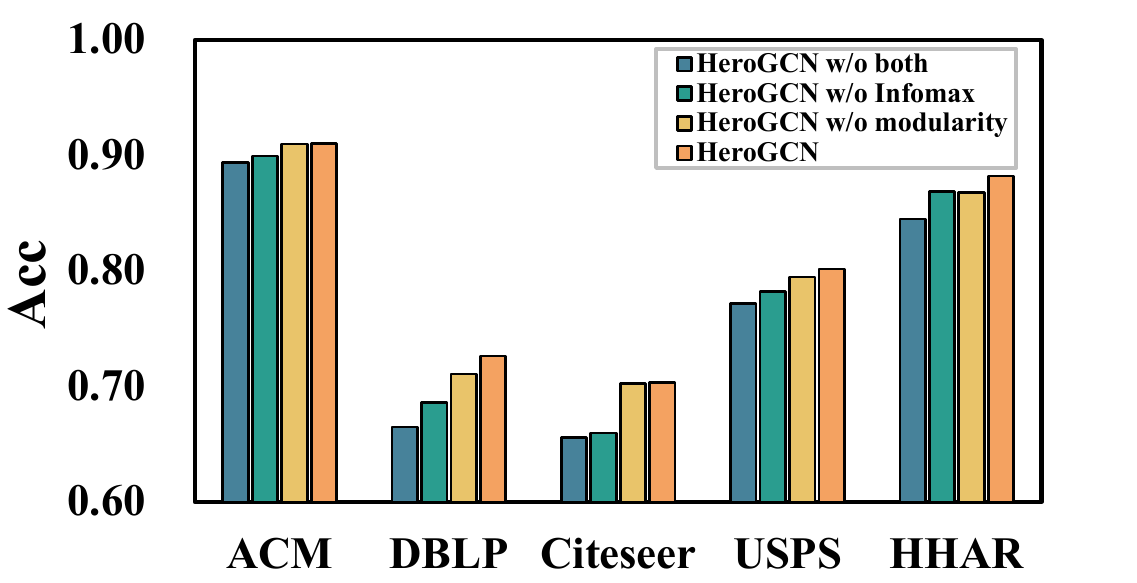} 
 \captionof{figure}{Ablation study results}
 \label{fig:Ablation}
 \end{minipage} 
  \vspace{-0.6cm}
 \end{figure}

The results in Fig.\ref{fig:Ablation} consistently demonstrate HeroGCN's outstanding performance. Even without modularity, our model excels, highlighting the effectiveness of the graph mutual infomax module. Notably, HeroGCN without infomax shows reduced effectiveness on the Citeseer dataset, likely due to its sparse graph nature and node classification into six categories. This complexity makes optimizing clustering based on the adjacency matrix challenging. Subsequent experiments reveal that clustering results on Citeseer already exhibit a modularity of 0.2994, posing difficulties in achieving further improvements through explicit graph structure supervision.

\section{Conclusion}\label{conclusion}
In this paper, we propose HeroGCN to leverage higher-order graph structural information for clustering. HeroGCN combines GCN and encoder to create an attribute-enriched graph convolutional network (AGCN) for hybrid representations. We use a graph mutual infomax module to capture the higher-order structural information and integrate modularity into the trinary self-supervised module for better supervision. Comparative experiments validate our method's superiority. In our future work, we will explore the adaptive fusion of GCN and encoder for further performance enhancement.
\\\\
\noindent\textbf{Acknowledgement.}\quad This research is supported in part by the National Science Foundation of China (No. 62302469), the Natural Science Foundation of Shandong Province (ZR2023QF100, ZR2022QF050), and the General Project for Undergraduate Education and Teaching Research at OUC (2023JY016).
\bibliographystyle{splncs04}
\bibliography{dasfaa}

\end{document}